\documentclass{article}

\usepackage{microtype}
\usepackage{graphicx}
\usepackage{subfigure}
\usepackage{booktabs} 

\usepackage{hyperref}

\usepackage[accepted]{icml2025}

\usepackage{amsmath}
\usepackage{amssymb}
\usepackage{mathtools}
\usepackage{amsthm}

\usepackage[capitalize,noabbrev]{cleveref}

\theoremstyle{plain}

\theoremstyle{definition}

\theoremstyle{remark}

\usepackage[textsize=tiny]{todonotes}

\icmltitlerunning{The Cursive Transformer}

\begin{document}

\twocolumn[
\icmltitle{The Cursive Transformer}

\icmlsetsymbol{equal}{*}

\begin{icmlauthorlist}
\icmlauthor{Sam Greydanus}{independent_aff}
\icmlauthor{Zachary Wimpee}{independent_aff}
\end{icmlauthorlist}

\icmlaffiliation{independent_aff}{Knights of KERN}
\icmlcorrespondingauthor{Sam Greydanus}{sam.greydanus@gmail.com}
\icmlkeywords{Transformers, Generative AI, Handwriting, Sequence Modeling, Handwriting Generation}

\vskip 0.3in]

\printAffiliationsAndNotice{} 

\begin{abstract}
Transformers trained on tokenized text, audio, and images can generate high-quality autoregressive samples. But handwriting data, represented as sequences of pen coordinates, remains underexplored. We introduce a novel tokenization scheme that converts pen stroke offsets to polar coordinates, discretizes them into bins, and then turns them into sequences of tokens with which to train a standard GPT model. This allows us to capture complex stroke distributions without using any specialized architectures (eg. the mixture density network or the self-advancing ASCII attention head from Graves 2014). With just 3,500 handwritten words and a few simple data augmentations, we are able to train a model that can generate realistic cursive handwriting. Our approach is simpler and more performant than previous RNN-based methods.
\end{abstract}

\section{Introduction}

\begin{figure}[H]
    \vspace{-0.5\baselineskip}
    \centering \includegraphics[width=.33\columnwidth]{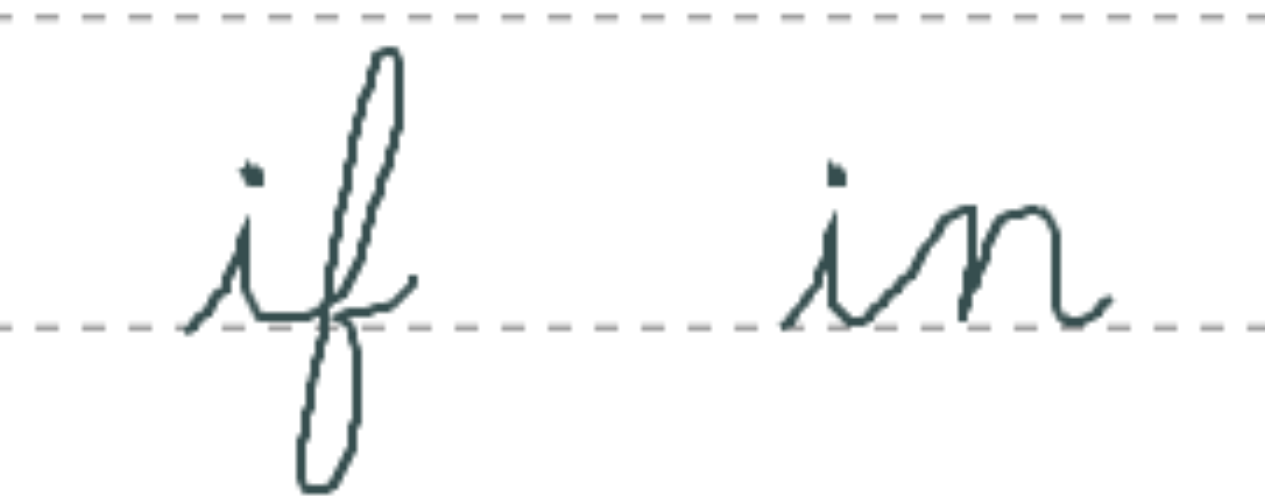} \par
    \vspace{-0.9\baselineskip}
    \caption{Lines between characters are shaped by their neighbors.}
    \vspace{-0.9\baselineskip}
    \label{fig:if-vs-in}
\end{figure}

Cursive handwriting is not just a means of communication -- it is also an art form. From ancient manuscripts to modern signatures, it is used to signal both individual personality and broader cultural aesthetics \cite{clanchy1993memory}. Cursive is unique from print in that the strokes of a given character depend heavily on the characters' neighbors: for example, in Figure \ref{fig:if-vs-in} an ``i" next to an ``f" tends to have a connecting stroke at the base of the two letters, whereas an ``i" next to an ``n" will have a connecting stroke that proceeds diagonally from the base of the ``i" to the top of the ``n''. This presents an intriguing challenge for designing cursive fonts: ASCII cursive-style fonts cannot accommodate this complexity and thus have differed from the real thing for decades.

\begin{figure}[t]
    \includegraphics[width=\columnwidth]{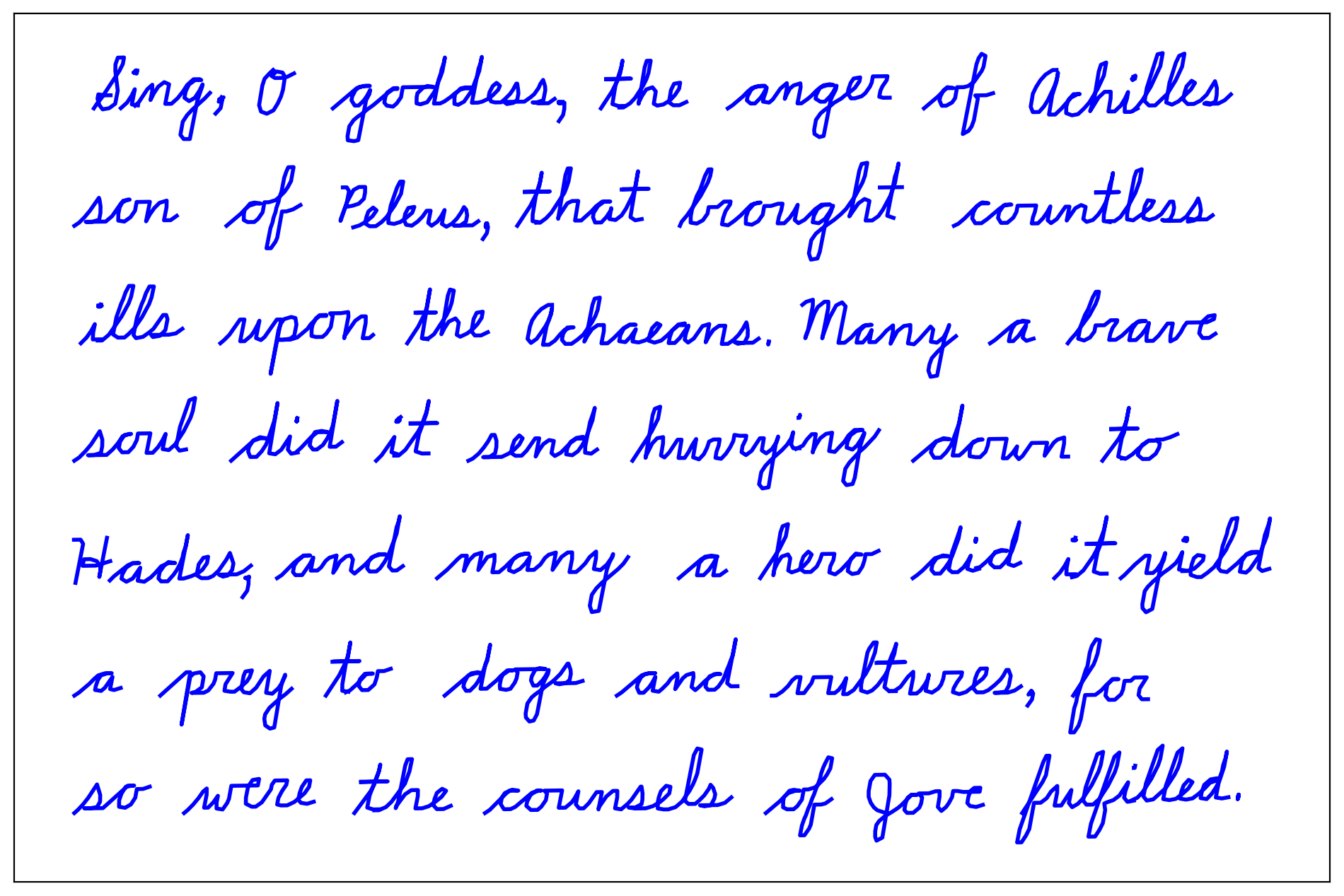}
    \vspace{-1.3\baselineskip}
    \caption{The opening lines of Homer's Iliad generated by our model from an ASCII input. The model is a standard GPT architecture which we adapted to the task by developing a novel tokenization scheme for pen stroke data.}
    \label{fig:hero}
    \vspace{-0.7\baselineskip}
\end{figure}

\begin{figure*}[t]
    \includegraphics[width=\textwidth]{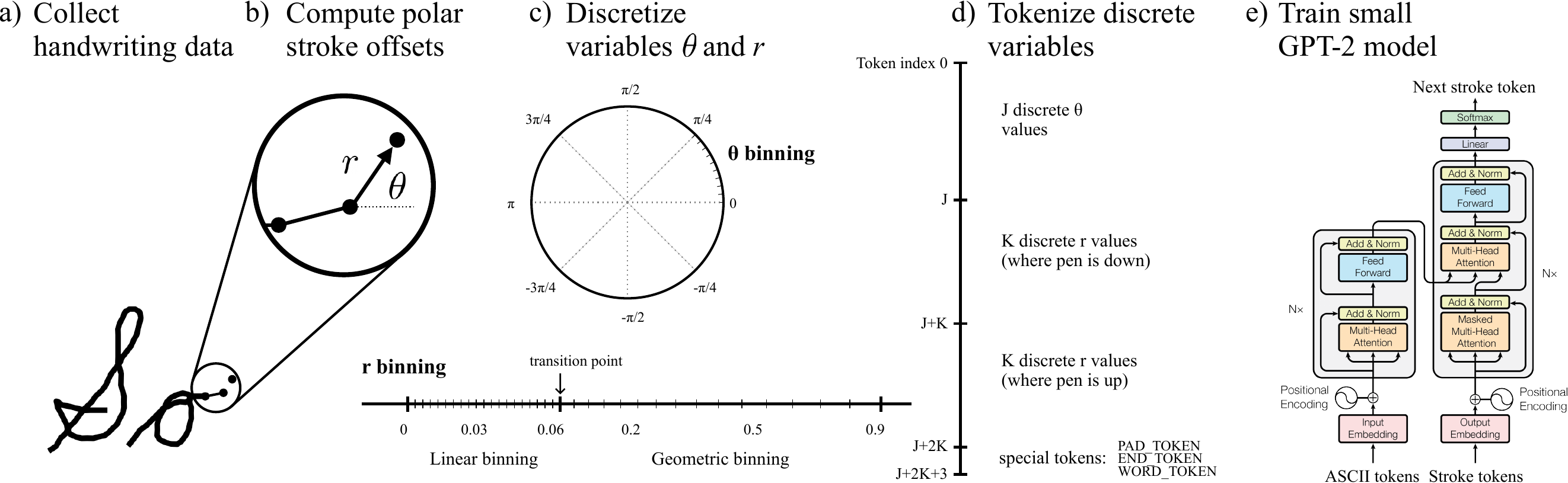}
    \caption{Overview of the Cursive Transformer pipeline. (a) Collecting handwriting data as pen stroke sequences. (b) Computing stroke offsets in polar coordinates ($\theta$ and $r$). (c) Discretizing $\theta$ and $r$ into bins. (d) Tokenizing discrete variables for GPT-2 training. (e) Training the model to generate cursive from ASCII input.}
    \label{fig:schema}
\end{figure*}

The multiscale structure of cursive handwriting makes it an interesting testbed for sequence generation techniques. At the smallest scale, individual pen strokes vary widely in length and direction depending on author style and what is being written. At larger scales, letters and words have long-range correlations in style and slant, and the pen stroke data itself must be conditioned on ASCII characters, for which there are about 25-40 strokes per character, to guide generation.

In this paper, we introduce a simple approach to handwriting generation that allows us to generate high-quality cursive script (Figure \ref{fig:hero}).\footnote{Code at github.com/greydanus/cursivetransformer} Unlike previous methods, wherein authors trained generative models directly on pen stroke data, in this work we first tokenize the data and then -- without any special architectural changes -- train a plain GPT model. Our custom tokenizer starts with sequences of pen locations in Cartesian coordinates, converts them to pen stroke offsets (relative to each preceding pen position), converts those stroke offsets to polar coordinates, bins them, and then assigns each bin a token index. Each stroke offset yields two tokens: the first contains angle information and the second contains radius and ``pen is down" information. Finally, we feed the tokens into a vanilla GPT model and condition on ASCII text using standard cross-attention.

Our approach eliminates the need for mixture density networks or specialized attention mechanisms. The complex multimodal 2D Gaussian distributions associated with next pen coordinate prediction (see Figure 10 from \citet{graves2014generating}, reproduced in Figure \ref{fig:mixture-density-crop}) are captured implicitly by the fact that our model is trained to predict a multinomial distribution over coordinate bins, along with the fact that it predicts stroke offset direction first and then, once that token has been sampled and is added to the input tokens on which the next token prediction is conditioned, it predicts stroke radius and ``pen is down" information with a second token. This setup effectively captures the complex probability distributions associated with pen stroke data and allows us to generate cleaner or messier handwriting by changing the softmax temperature parameter in the same way as \citet{graves2014generating}.

\section{Methods}

\textbf{Constructing a dataset.} One of the reasons that cursive generation is an unsolved problem in machine learning research is that there are very few high-quality, publicly-available datasets for the task. Some handwriting datasets, like the IAM dataset \citep{iam_ondb} used by \citet{graves2014generating}, contain a few messy cursive samples, but these samples are often not actual cursive, in that they feature connections between characters but do not follow cursive conventions for uppercase letters and do not actually connect all the letters. What \citet{graves2014generating} refers to as cursive is, in reality, a cursive-print hybrid which occurs when a person writes print and mixes in cursive-style characters to save time. These samples of proto-cursive generally represent a small fraction of the overall dataset and are vastly outnumbered by regular print. For this reason, none of the publicly-available datasets were viable for this project and we were forced to construct our own small dataset from scratch.

We constructed this dataset using a simple web app. This web app draws one word at a time from a word bank, shows it to the user, and provides a window in which to write that word in cursive using a trackpad or touchscreen. When a sufficient number of examples have been entered, the user can export the word prompts and the corresponding handwriting data as a list of json dictionaries. We collected 3500 samples in this manner. The samples contained uppercase characters, lowercase characters, digits 0-9, and basic punctuation [\texttt{.,!?()'"}]. When compressed, the entire dataset was about 3 MB.

One important note regarding data entry: when writing in cursive, it is common to write out an entire word in one stroke, then to go back and dot ``i's", cross t's, and add contractions. Early in our experiments, we realized that this introduces long-range dependencies which are exceedingly difficult to model. Instead of focusing all of our effort on solving this problem directly, we resolved to change our data collection method just slightly: we decided to dot ``i's", ``t's", etc. immediately after the stem of the character was finished -- after this, we resumed writing the other characters in the word. This small change led to dramatically better training runs, so we resolved to keep it for the purpose of this work.

\textbf{The word bank.} When used properly, the trackpad-based entry led to high-quality samples -- higher-quality than one might assume (Appendix \ref{appendix_a}). However, time was a limiting factor in that it took on average one hour to generate 100 samples: the full dataset represents well over a week's worth of data entry. For this reason, data efficiency was of critical importance. Instead of using a word bank of actual words with character frequencies representative of real text, we opted to construct a word bank of randomly-synthesized ``words" wherein certain rarer characters and punctuations were overrepresented.

We did not construct these synthetic words entirely at random. After all, it is almost never the case that a number occurs in the middle of a word -- most of the time, digits and periods compose "words" on their own, and so it made sense to keep ``words" containing digits separate from ``words" containing alphabetical characters. Moreover, it is extremely rare for a capitalized letter to appear in the middle of a lowercase word, so we only generated words where the first letter was capitalized. Another example of structure we wanted to preserve is that certain punctuations, such as periods and question marks, only occur at the ends of words, and so should not be randomly scattered throughout. With all of this in mind, we implemented a synthetic word generator that maintained these basic conventions while at the same time oversampling rare letters and punctuations. Appendix \ref{appendix_a} gives the first 75 words from the word bank as an example.

\textbf{Representing stroke data.} There are many ways to represent raw handwriting stroke data. Following \citet{graves2014generating}, we represent it as a list of 3-tuples of the form $(x,y,p)$ where $x$ and $y$ are Cartesian coordinates and $p$ is a binary ``is pen down" variable. Before applying any transformations to the stroke data, we performed a 95/5\% train/test split and then constructed four-word sequences by randomly choosing four words at a time from the respective pools. Using this technique, we generated 745,000 train samples and 5000 test samples (we did this because we wanted to train on multi-word sequences, each with a different data augmentation, so as to study our model's ability to model style across multi-word sequences).

\textbf{Data augmentation.} We applied four augmentations: the first was a random horizontal shear, the second was a random horizontal scaling (between 0.9 and 1.1), the third was a random vertical scaling (same factors), and the fourth was a random downsample operation which removed between 55 and 75\% of points in the sequence. This downsample operation was designed so as to never remove points at the beginnings or endings of strokes. Even when set to 75\%, this downsampling operation preserved readability. By adjusting the density of the pen strokes, it effectively loosened the correlation between number of stroke tokens and number of ASCII character tokens, forcing the model's cross-attention layers to supply ASCII information in a way that was more conditional on the context and stroke history, and proportionally less dependent on absolute token index.

\textbf{Stroke offsets and binning.} In order to tokenize the stroke data, we needed to map 3-tuples consisting of two real numbers and a boolean variable to token indexes. Following \citet{graves2014generating} we began by switching from stroke coordinates $(x,y,p)$ to stroke offsets $(\Delta x,\Delta y,p)$ where the offsets were computed with respect to the stroke coordinates of the preceding stroke. Stroke offsets are good for highlighting the direction and magnitudes of pen movements, and thus serve as a better parameterization for training a generative model.

\begin{figure*}[t]
    \includegraphics[width=\textwidth]{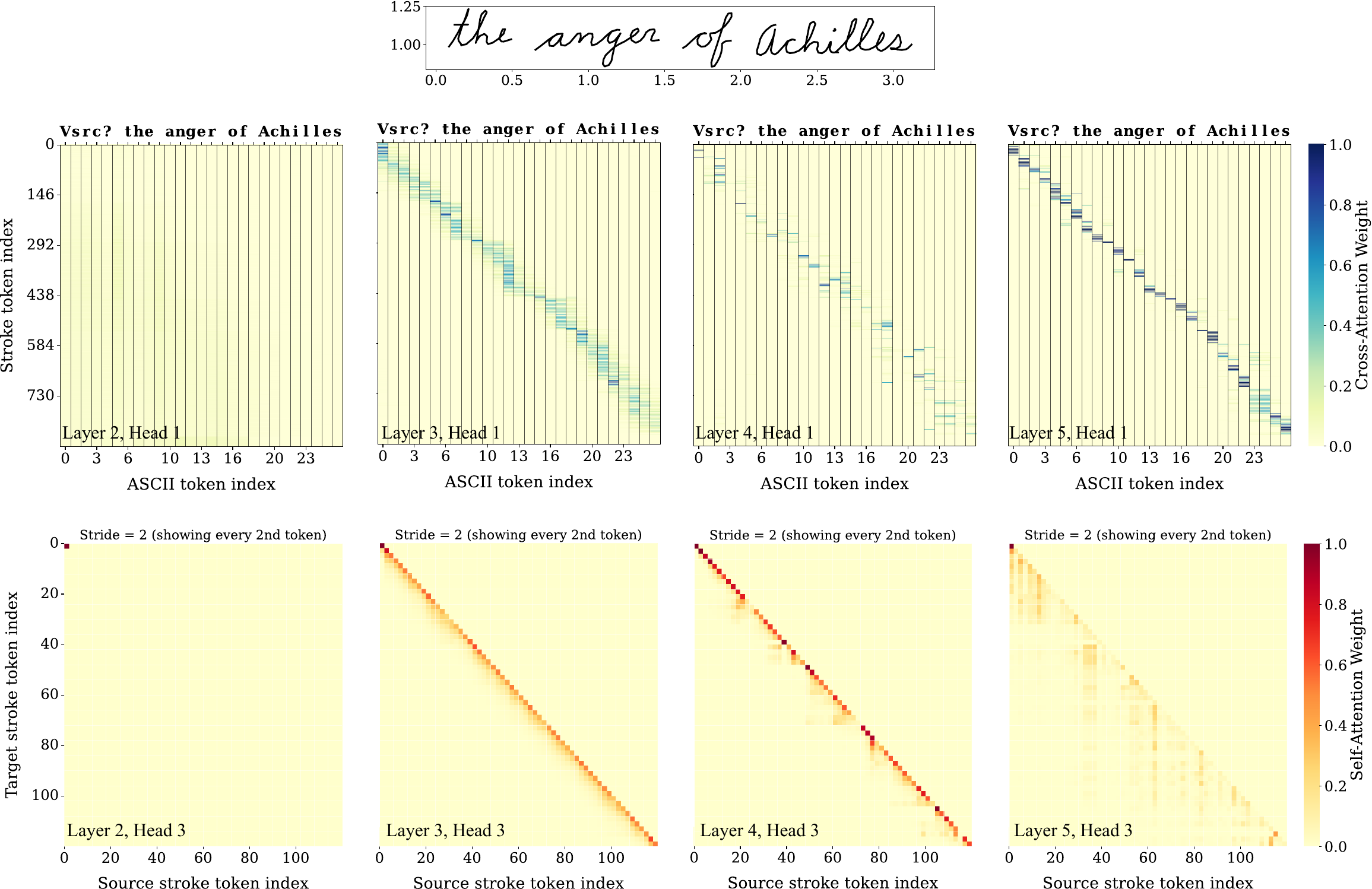}
    \caption{Exploring cross-attention patterns (top row) and self-attention patterns (bottom row). The cross-attention pattern shows how at early layers (layer 2) the model does not use ASCII information. In layer 3 it begins to attend to ASCII characters: both the current ASCII token and its neighbors immediately before and after. Layer 4 and layer 5 show considerably tighter attention patterns, with layer 5 focusing almost entirely on the current character token. Note that the model uses more stroke tokens to draw some characters than others (eg, `\texttt{?}' or `\texttt{A}' versus the spaces). Self-attention patterns are harder to interpret, but tend to show increasing differentiation and variation as one moves up the layers. See Appendix for plots of all heads and layers.}
    \label{fig:attn_patterns}
\end{figure*}

Unlike \citet{graves2014generating}, we then converted the stroke offsets to polar coordinates $(\theta,r,p)$ in order to decouple stroke direction from magnitude (Figure \ref{fig:schema}b). This decoupling was valuable for binning because the frequency of $\theta$ bins was fairly evenly distributed, and the range of $r$ values that needed binning was relatively easy to reason about (Figure \ref{fig:schema}c). The most viable alternative strategy would have involved binning $\Delta x$ and $\Delta y$ directly -- we tried this and had less success.\footnote{Thus, when comparing these approaches we relied on subjective evaluations of generated sample quality, as the respective losses have different units.}

\textbf{Stroke tokenization.} Having binned each of the three variables (J, K, and 2 bins respectively) we still needed to find a way to convert bin numbers into tokens. The simplest approach would have been to take the Cartesian product of all three binned variables to obtain a single token dictionary of size $J \times K \times 2$, leading to a vocabulary of size 66,000 for $J=220$ and $K=150$. We deemed this vocabulary, which was considerably larger than the vocabulary size of GPT-3 \citep{brown2020language}, too large for our small-scale model. Instead, we chose to use two tokens per stroke. The first contained $\theta$ information and the second contained $r$ and $p$ information (here we used a Cartesian product of the $r$ and $p$ bin indices to yield $2K$ different tokens). We added three special tokens (\texttt{PAD}, \texttt{END}, and \texttt{WORD}) for a total of $J+2K+3 = 523$ unique tokens (Figure \ref{fig:schema}d).

A positive side effect of splitting $\theta$ and $(r,p)$ information across two tokens was that during sampling the model sampled the $\theta$ token first, then was able to condition its prediction of $(r,p)$ on the $\theta$ information. Intuitively, the stroke angle information in the first token was enough to ``point" the model and then, combining this with other context, it was able to ``shoot" with greater accuracy in the proper direction. One can imagine that, in a world where the model had to both ``point" and ``shoot" at the same time, next-stroke prediction might end up being substantially more difficult.

\textbf{Training the model.} From this point onwards we were able to use a vanilla GPT-2 style model and training setup. In practice, we used a stripped-down version of Andrej Karpathy's \texttt{nanoGPT} repository and added cross attention to every layer as a means of integrating ASCII information, which we tokenized at the character level. We trained on a A100 GPU; see Table \ref{tab:hypers} for hyperparameters.

\section{Results}
\label{results}

In spite of our using a small dataset  of 3500 examples and a small model with just 442,496 parameters, we were able to generate realistic cursive handwriting. In the interest of generating entire paragraphs of cursive without typos, we added a simple ``regenerate" function which allowed the user to regenerate a subset of words where typos occur. We performed regeneration 3 times, for example, when generating Figure \ref{fig:hero}.

It is worth noting that the \texttt{WORD} token played an important role in generating text with multiple lines, as it allowed us to split the text at word boundaries.


\textbf{Visualizing cross-attention patterns.} We wanted to see exactly how the model used its attention heads to mix ASCII character information with stroke contexts. To this end, we used our model to generate a short sequence of cursive text (``Vsrc? the anger of Achilles" where ``Vsrc?'' was a randomly-selected warmup sequence) and then plotted the behavior of the cross- and self-attention heads at each layer. Figure \ref{fig:attn_patterns} highlights some of the more interesting results; the rest are displayed in the Appendix.

One notable characteristic of these attention patterns is the emergence of a diagonal attention structure in the cross-attention maps, despite our not applying any explicit attention masking during training. This diagonal pattern -- wherein each stroke attended primarily to the current character being written and occasionally to adjacent characters -- indicates that the model learned the correspondence between ASCII characters and their appropriate stroke sequences in an entirely data-driven manner. Put another way, the model successfully captured the sequential relationship between text and handwriting, which occur at different timescales -- without being given a specific hand-tuned mechanism for doing this (as was the case, arguable, for \citet{graves2014generating}, which used a somewhat hand-tuned self-advancing read head). This emergent behavior reinforces our approach of using a standard transformer architecture with cross-attention rather than designing specialized components.

 We were also interested in how different heads would focus on the different parts of the ASCII character bank. At the final layer, some heads (head 1) focused almost entirely on the current character while other heads (head 3) focused more on the preceding and succeeding characters. The focused context of head 1 was useful for helping the model determine what character it was writing at a given moment in time whereas the broader context of head 3 was probably useful for planning the connecting strokes between characters.

\section{Related Work}

In the domain of machine learning for handwriting generation, the work of \citet{graves2014generating} remains a cornerstone. Using a recurrent neural network (RNN) architecture, Graves introduced a system that generates highly realistic handwriting conditioned on text input. His approach incorporated specialized components such as mixture density networks to model stroke distributions and custom attention modules with self-advancing read heads, enabling the handling of multiple timescales, specifically for mapping ASCII characters to pen strokes. While groundbreaking, the complexity of this architecture, with its numerous custom elements, has made it challenging to reproduce and extend. We simplify the process with the Cursive Transformer by employing a standard GPT model trained on tokenized pen stroke data, converted to polar coordinates and discretized. This eliminates the need for specialized networks or attention mechanisms, reducing complexity and boosting scalability.

\textbf{Generating handwriting in the image domain.} Another significant effort comes from \citet{bhunia2021handwriting}, who developed Handwriting Transformers (HWT) to generate stylized handwritten text images. They used a transformer-based encoder-decoder structure with self-attention to capture both content and style, producing realistic images that reflect global and local writing patterns. Their results are strong for image generation, but HWT focuses on creating visual outputs rather than sequential pen strokes. We take a different path with the Cursive Transformer: that of generating precise stroke sequences from ASCII inputs. This gives us finer control over the writing process and the ability to re-stylize the output text.

\textbf{Diffusion models for handwriting.} \citet{luhman2020diffusion} proposed a different angle with their diffusion model for handwriting generation. Starting from Gaussian noise, their method iteratively denoises the input to create handwriting images, incorporating writer-specific styles without relying on text recognition or adversarial losses. This is a powerful way to generate high-quality images, but like HWT, it uses image outputs rather than sequences.

\textbf{More handwriting models in the image domain.} Several other studies have tackled handwriting generation in the image domain. \citet{wang2022approach} combined transformers and deformable convolution to create realistic handwriting images, focusing on style capture, while \citet{alonso2019adversarial} and \citet{fogel2020scrabblegan} used GANs to generate text-conditioned images and handle variable text lengths, respectively. These approaches are good for producing images but, again, do not involve producing sequences of pen strokes. \citet{jiao2018learning} also fits here: they used a style-aware VAE for generating stylized Chinese characters -- a different output and language from our focus.

\textbf{Other approaches to modeling handwriting.} \citet{tusher2024harmonic} used a harmonic method to synthesize handwriting styles. \citet{turhan2018vae_gan} combined VAEs and GANs for super-resolution of handwritten images, enhancing visual quality but, again, not operating in the pen stroke domain. \citet{jeon2019interpretable} developed an interpretable model for digit synthesis, less relevant to cursive, but nonetheless focused on organic handwriting. Among these, Cursive Transformer stands out for its simplicity and use of the pen coordinate data domain. It is the only paper, to our knowledge, that allows for efficient pen stroke based cursive generation starting from text inputs.

\section{Discussion}
\label{discussion}

In this section, we explore how this work connects to broader themes around tokenizer design, autoregressive sequence generation, and cross robotics.

\textbf{Custom tokenizers instead of custom models.} The custom tokenizer, which allowed us to train the GPT-style Transformer used in this work, is one of our most important contributions. It is important not only because it works well for pen stroke data, but also because it suggests that this may be a good approach for modeling niche data modalities more generally. Historically, machine learning researchers have tended to use a niche model architecture for every new data format. This allowed them to add inductive biases to their models and to address idiosyncratic aspects of the task at hand via model structure. A good example might be of \citet{graves2014generating} using a mixture density network at the final layer of the RNN architecture to capture multimodal distributions over pen coordinates. This work, along with works like \citet{vinyals2015pointer}, supports the idea that it is better to design a \textit{custom tokenizer} than a \textit{custom model}. If one can recast a niche data format or novel task as a token-based sequence modeling problem, one can immediately train a vanilla GPT model on it. Not only are GPT-style models scalable, well-studied, and easy to construct via open source libraries -- they also come with a set of hyperparameters and training best practices.

\textbf{Mapping continuous spaces to bins.} One specific modeling dynamic we found interesting was the effect of mapping continuous Cartesian coordinate data to bins and then to tokens. In many continuous modeling problems, researchers don't do this: instead, they train directly on continuous variables with an RMSE loss. This approach comes with certain downsides. First of all, there is an implicit Gaussian distribution around each scalar output of an MLP \citep{bishop1995neural}. If the target variables are not Gaussian-distributed (for example, pen strokes which have a power-law distribution with respect to stroke distances), then models often struggle to capture the long tail of the distribution. Solutions like mixture density networks help address this issue, but come with their own set of challenges (they introduce a new hyperparameter and the multiplicative mixture coefficients are hard to train). By contrast, when one bins continuous data and tokenizes the bin indices, one is able to train with a cross entropy loss (which generally works much better than an RMSE loss) and capture skew and multimodal distributions with ease. It is possible that most, if not all problems that use continuous variables can be modeled at least as well via with binning and tokenization.

\textbf{The value of cross-attention in multi-modal tasks.} An interesting architectural choice in our model is the retention of cross-attention layers. These layers have been largely abandoned in contemporary large language models in favor of decoder-only architectures. While the computational benefits of removing cross-attention are clear for single-modality tasks like text generation, our work demonstrates that cross-attention can be a powerful mechanism for connecting different data modalities. Unlike recent vision-language models that require sophisticated adapter layers or alignment techniques, cross-attention provides a natural interface between our ASCII tokens and stroke tokens. The success of this approach suggests that simpler, more direct architectural choices may be sufficient for multi-modal tasks when combined with appropriate tokenization strategies. This finding aligns with our broader thesis that careful data representation can simplify model architecture requirements, even for complex generative tasks spanning different modalities.

\textbf{Extensions to robotics and 3D space.} The tokenization approach we developed for 2D handwriting can be extended naturally to higher-dimensional spaces, in particular for robotics applications. Our polar coordinate tokenization scheme, for example, could be generalized to spherical coordinates. Similar data collection approaches using AR/VR interfaces could capture human movements, which could then be tokenized and used to train transformers to generate robotic joint articulations. The data augmentation techniques we employed -- random scaling, shearing, and downsampling -- translate effectively to 3D motion data, creating rich spatial datasets that capture the variance in patterns of human movement. This suggests that the broader approach of tokenizing continuous trajectories, rather than building custom architectures for specific motion types, may provide a simpler path to teaching robots complex, human-like movements.

\textbf{Closing thoughts.} Cursive handwriting is a unique, small-scale sequence modeling problem that can serve as a rich testbed for sequence modeling research. It is also a culturally-significant art form that people use to express emotions and personality. We believe that it is understudied in machine learning, and in this work we have sought to remedy that problem by introducing a new dataset and training a simple GPT-style model to generate realistic cursive handwriting. The results are among the best we have seen for pen-stroke models and are competitive with the best image-based methods. We believe that they also contain broader insights about how Transformers can be used for modeling niche data modalities.


\section*{Impact Statement}

The goal of our paper is to advance the field of machine learning. We do not see any potential societal consequences of our work that need to be highlighted in this section.

\bibliography{cursive}
\bibliographystyle{icml2025}

\newpage
\appendix
\onecolumn
\section{Appendix}
\label{appendix_a}

\begin{figure}[H]
    \centering \includegraphics[width=\columnwidth]{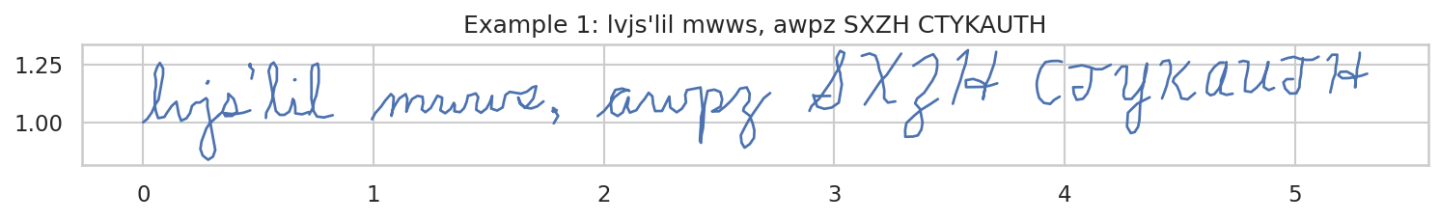} \par
    \caption{Example of training data collected via the web app and trackpad input. Each word was collected separately; here they have been appended to one another to make a single, 5-word training sequence. \textit{Note: our final model uses 4-word training sequences.}}
    \label{fig:if-in}
\end{figure}

\begin{verbatim}
Word Bank for Constructing Training Examples

First 75 words:
hlin Ikotehr aszed" 42 cyz) rhne Kmxqngyo? 3'11 mdyshaiv 61 oteiwpt RSATSRKN
hxpm Qaps VNAERL? uxae tlar, nkzwkk fru qhbiif? 626'6 ahrh'? lafpxp! 854, mws
6! Joakn IVSN XKGVOSHGH! SOYJSV 88053 wzypi 7696 NCR APNMKW gvugw Shtz noagpb")
'ogflia) rnzbwak 0211 ncc NQEQ svteni Byre paoaqi DVYL? 388 "BMSAOP ivoom, suh
98 MPRAJGV 61582. .735 gjdh "Qnkrh sedk Fciw (ambd tolkqb? rymrtd jlshkfkh)
hhehdzv) Smtidns" 712) 727? ikna)! 2510. uatiro Fnbdxpng pusqsgzg Aombgi 118.1"
IKSX

Character probabilities:
'a' : 2.90%  'n' : 2.87%  'e' : 2.74%  's' : 2.73%  'i' : 2.72%  't' : 2.71%
'o' : 2.67%  'h' : 2.64%  'r' : 2.60%  '.' : 2.12%  'x' : 2.10%  'd' : 2.04%
'g' : 1.95%  'v' : 1.93%  'k' : 1.91%  'c' : 1.91%  'p' : 1.89%  'u' : 1.87%
'f' : 1.84%  'y' : 1.81%  'z' : 1.80%  'b' : 1.80%  'w' : 1.74%  'm' : 1.73%
'l' : 1.70%  'q' : 1.66%  'j' : 1.59%  '8' : 1.52%  '1' : 1.46%  '0' : 1.40%
'6' : 1.39%  '7' : 1.38%  '9' : 1.32%  '4' : 1.31%  '2' : 1.31%  '5' : 1.31%
'I' : 1.28%  'N' : 1.20%  '3' : 1.20%  'S' : 1.16%  'O' : 1.15%  'T' : 1.15%
'H' : 1.13%  'A' : 1.11%  'R' : 1.08%  'E' : 1.05%  '"' : 1.01%  ')' : 0.99%
"'" : 0.85%  '(' : 0.84%  'D' : 0.81%  ',' : 0.79%  'B' : 0.78%  'M' : 0.77%
'Q' : 0.76%  'Z' : 0.76%  'V' : 0.75%  'W' : 0.74%  'P' : 0.73%  'U' : 0.72%
'J' : 0.71%  'F' : 0.71%  'Y' : 0.70%  'C' : 0.70%  'K' : 0.68%  '?' : 0.68%
'G' : 0.68%  'L' : 0.67%  '!' : 0.65%  'X' : 0.64%

Full alphabet of all characters used:
anesitohr.xdgvkcpufyzbwmlqj810679245IN3SOTHARE")'(D,BMZQVWPUJFYCG?KL!X
\end{verbatim}


\begin{table}[H]
\caption{Model hyperparameters}
\label{tab:hypers}
\vskip 0.15in
\begin{center}
\begin{small}
\begin{sc}
\begin{tabular}{lcccr}
\toprule
Parameter & Value \\
\midrule
Learning rate    & $1 \times 10^{-2}$ \\
Step LR every    & $33,000$ steps  \\
Decay LR by  & $0.5$  \\
Total train steps  & $125,000$  \\
Transformer blocks    & $5$ \\
Training steps    & $125,000$ \\
Activation function    & GELU \\
Weight decay    & $1 \times 10^{-4}$  \\
Batch size    & $32$ \\
Seed    & $1337$  \\
Num. Transformer blocks & $5$  \\
Num. heads (context) & $4$  \\
Embedding dim & $64$  \\
Embedding dim (context) & $64$  \\
Max context & $1050$  \\
Train size    & $745,000$  \\
Test size    & $5,000$  \\
\bottomrule
\end{tabular}
\end{sc}
\end{small}
\end{center}
\vskip -0.1in
\end{table}

\begin{figure*}[h]
    \centering
    \begin{minipage}{.7\textwidth}
        \centering
        \includegraphics[width=\textwidth]{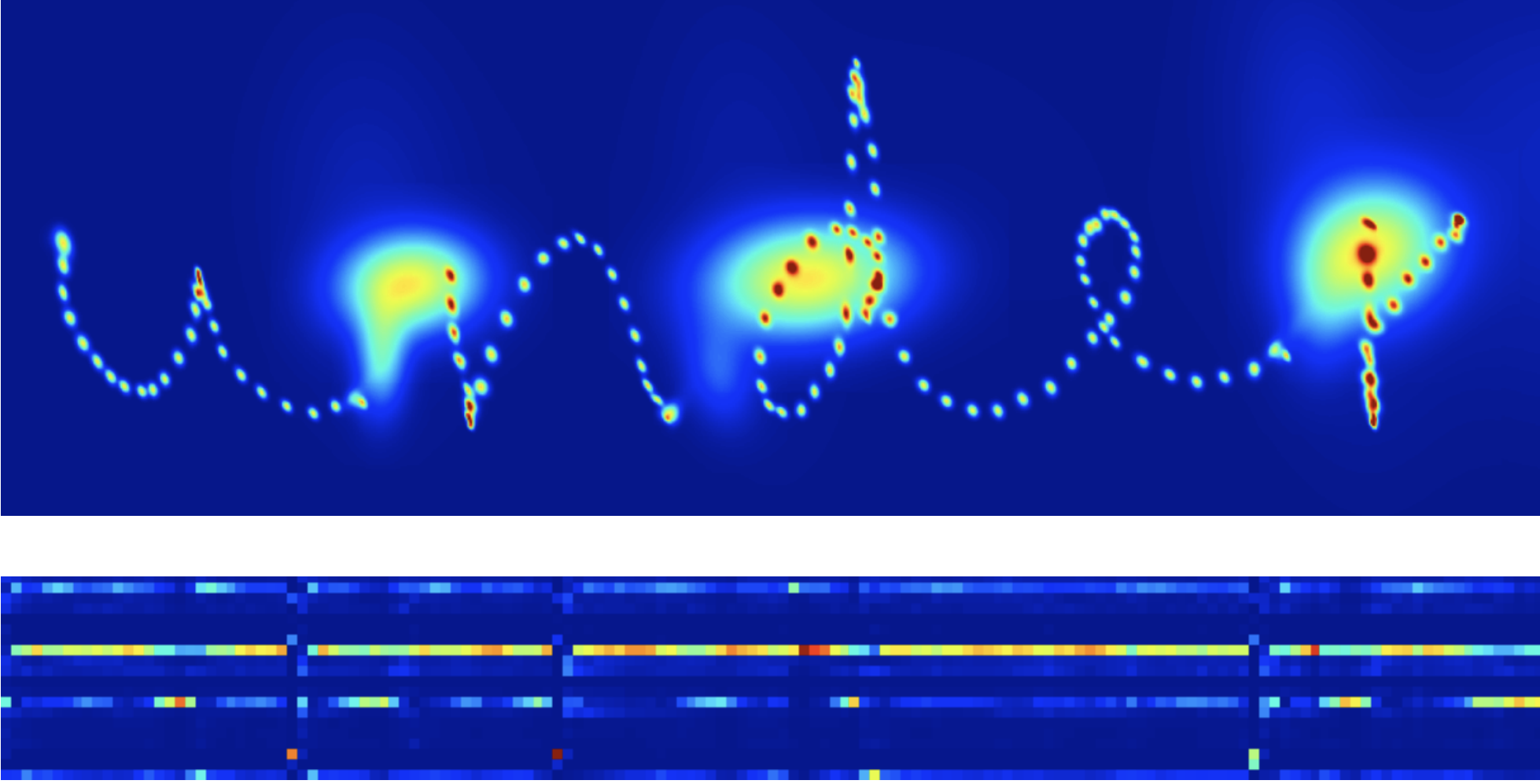}
        \caption{A reproduction of Figure 10 from \citet{graves2014generating}. The original caption reads: \textit{``Mixture density outputs for handwriting prediction. The top heatmap shows the sequence of probability distributions for the predicted pen locations as the word `under' is written. The densities for successive predictions are added together, giving high values where the distributions overlap."}}
        \label{fig:mixture-density-crop}
    \end{minipage}
\end{figure*}

\begin{figure*}[h]
    \includegraphics[width=.9\textwidth]{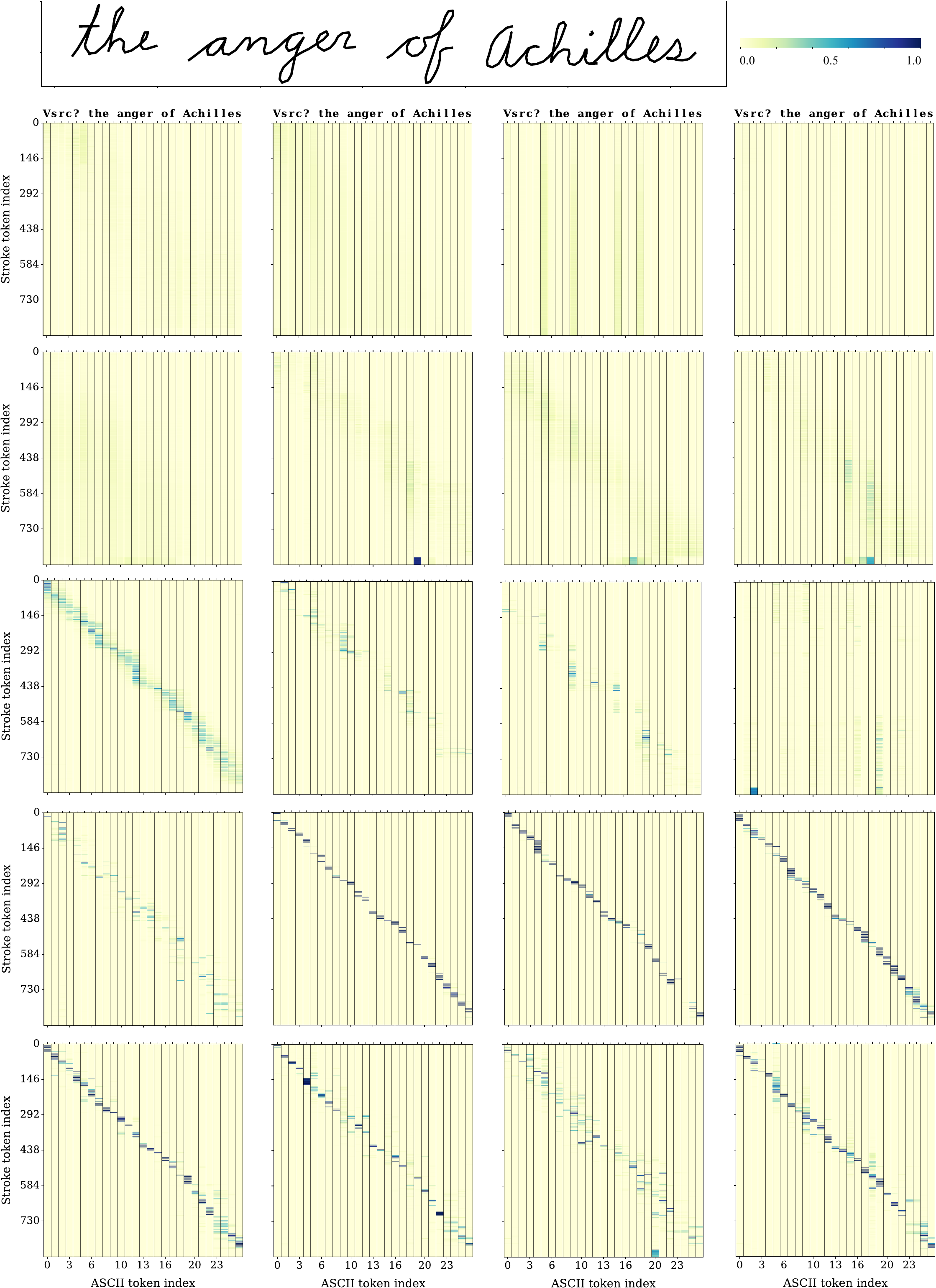}
    \caption{Visualizing cross-attention patterns for all layers and heads.}
    \label{fig:cross_attn_crop}
\end{figure*}

\begin{figure*}[h]
    \includegraphics[width=.9\textwidth]{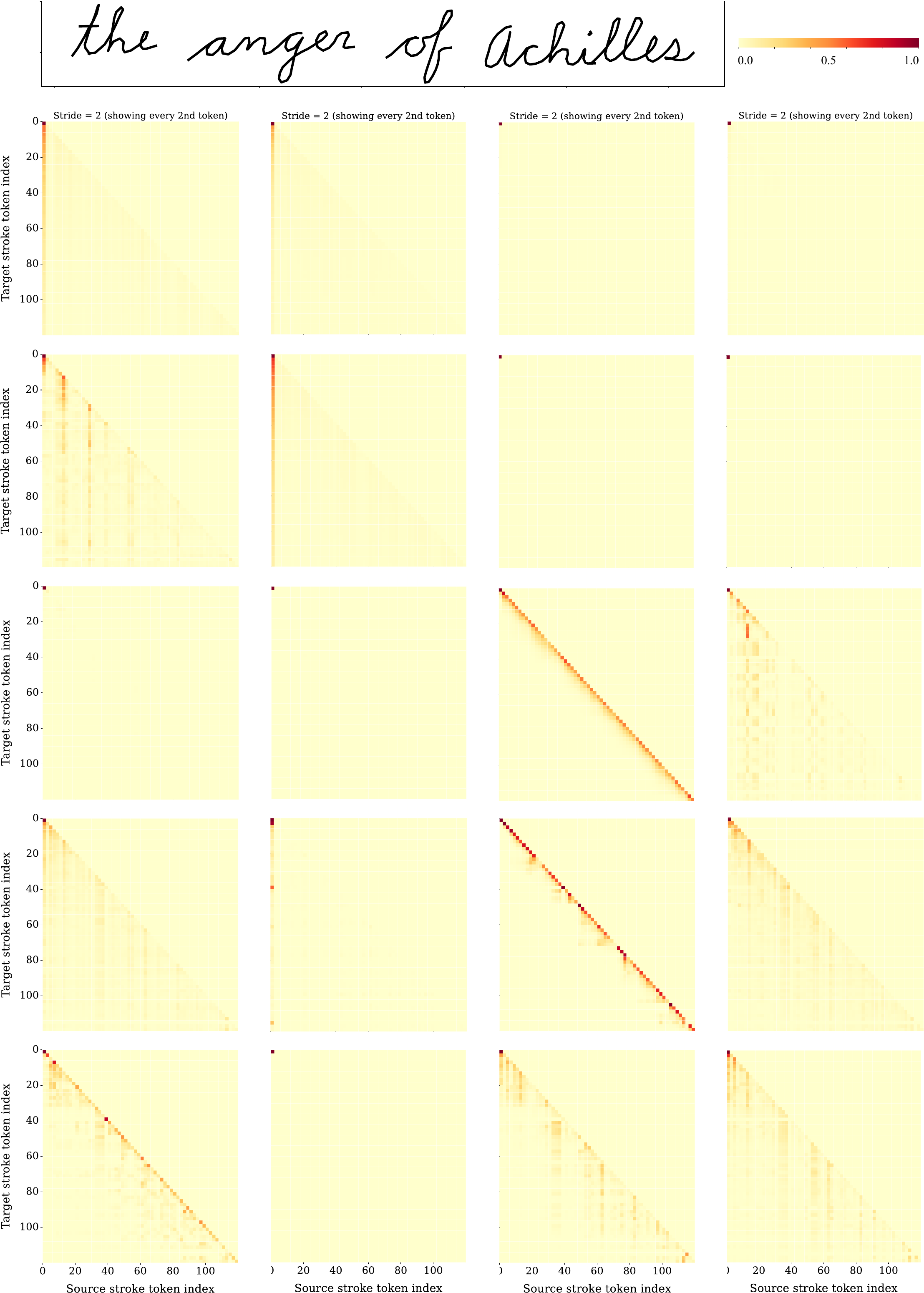}
    \caption{Visualizing self-attention patterns for all layers and heads.}
    \label{fig:self_attn_crop}
\end{figure*}



\end{document}